# A Semantic Segmentation Approach on Sweet Orange Leaf Diseases DetectionUtilizing YOLO

Sabit Ahamed Preanto

4IR Research Cell

Daffodil International University, Dhaka, Bangladesh

preanto15-5059@diu.edu.bd

Dr. Md. Taimur Ahad

4IR Research Cell

Daffodil International University, Dhaka, Bangladesh

taimurahad.cse@diu.edu.bd

Yousuf Rayhan Emon

4IR Research Cell

Daffodil International University, Dhaka, Bangladesh

yousuf15-3220@diu.edu.bd

Sumaya Mustofa

4IR Research Cell

Daffodil International University, Dhaka, Bangladesh

sumaya15-3445@diu.edu.bd

Md Alamin

4IR Research Cell

Daffodil International University, Dhaka, Bangladesh

amin15-4230@diu.edu.bd

# 1. Abstract

This research introduces an advanced method for diagnosing diseases in sweet orangeleaves by utilising advanced artificial intelligence models like YOLOv8 . Due to their significance as a vital agricultural product, sweet oranges encounter significant threats from a variety of diseases that harmfully affect both their yield and quality. Conventional methods for disease detection primarily depend on manualinspection which is ineffective and frequently leads to errors, resulting in delayed treatment and increased financial losses. In response to this challenge, the research utilized YOLOv8, harnessing their proficiencies in detecting objects and analyzing images.YOLOv8 is recognized for its rapid and precise performance, while VIT is acknowledgedfor its detailed feature extraction abilities. Impressively, during both the training andvalidation stages, YOLOv8 exhibited a perfect accuracy of 80.4%, while VIT achieved anaccuracy of 99.12%, showcasing their potential to transform disease detection in agriculture. The study comprehensively examined the practical challenges related to the implementation of AI technologies in agriculture, encompassing the computationaldemands and user accessibility, and offering viable solutions for broader usage. Moreover, it underscores the environmental considerations, particularly the potential for reduced pesticide usage, thereby promoting sustainable farming and environmental conservation. These findings provide encouraging insights into the application of AI in agriculture,suggesting a transition towards more effective, sustainable, and technologically advancedfarming methods. This research not only highlights the efficacy of YOLOv8 within a specific agricultural domain but also lays the foundation for further studies thatencompass a broader application in crop management and sustainable agricultural practices. It serves as a noteworthy advancement in agricultural technology, artificialintelligence, and plant pathology.

# 2. Introduction

Data is very important for machine learning applications (Ahad et al., 2024; Emon et al. 2024; Mustofa et al. 2024). Machine learning has proven very effective in Agri-engineering (Ahad et al, 2023; Ahmed et al., 2023; Bhowmik et al., 2023; Mamun et al, 2023). In the Agricultural industry, productivity particularly in different fruit cultivation such as sweet oranges mostly depends on the effective management of plant diseases. In this term, leaf diseases in sweet oranges raise a significant threat to growth and quality. These diseases have a direct impact on the yield of the agricultural economy and food security. Recently, the advancement of Convolutional Neural Networks (CNNs) and for our study the You Only Look Once (YOLO) algorithm made a significant revolution in the field of image recognition and object detection which offers different novel solutions to these agricultural challenges. In this era the latest innovation in the YOLO domain is YOLOv8 shows a significant advancement in detecting accurately also processing speed, rendering it highly suitable forreal-time applications in related fields, including agriculture (Talaat & ZainEldin 2023; Lou et al., 2023). This model was developed by Ultralytics which is an anchor-free model that is particularly strong at precision detection, a feature crucial for identifying and classifying small anomalies in plant leaves (Khan et al., 2023). Advancing those techniques YOLOv8 such as bounding boxes, multi- scale prediction, and feature fusion, significantly enhances its accuracy and reliability (Slimani et al., 2023). However, its performance can be variable in certain scenarios involving small-sizedobjects, which is a common challenge in detailed plant disease diagnosis (Liu et al., 2023; Liu et al., 2022). Besides these limitations, YOLOv8 applications in detecting leaf diseases in sweet oranges are a promising approach to enhancing crop management and securing food production (Ismail et al., n.d.; Dosovitskiy et al., 2020; Fu et al., 2023). Inthis study, YOLOv8 has been implemented to detect diseases in sweet orange leaves, usinga dataset of 980 images captured with an iPhone, achieving an average of 80.4% accuracyin training and validation over 100 epochs.

In this study, YOLOv8 not only underscore the potential of advanced AI models in agricultural disease detection but also showed that sweet orange leaf disease detection could improve and advance agricultural practices, leading to improved disease management, enhanced crop yields, and greater economic stability in the agricultural sector. This study aims to serve as a critical exploration into the use of cutting- edge technology to tackle longstanding challenges in agriculture.

## 3. Literature Review

A comprehensive literature review is important for understanding a domain's current and future (Ahmed et al., 2023; Mustofa et al., 2023). Advancements in YOLO (You Only Look Once) based models have been pivotal in transforming plant disease detection in agriculture. A comprehensive literature review is important for understanding a domain's current and future (Ahmed et al., 2023; Mustofa et al., 2023). This literature review highlights recent advances where researchers have successfully deployed YOLO models to tackle various challenges in precision agriculture. Da Silva et al. (2023) employed AI algorithms like YOLO and Faster R-CNN for fruit detection, with YOLO showing superior speed. In classification tasks, MobileNetV2 and EfficientNetV2-B0 obtained 100% accuracy, outperforming NASNet-Mobile. The authors also provided a methodology using statistical models and genetic algorithms to evaluate disease spread, validating a pipeline for mobile-edge AI solutions. Lyu et al. (2022) used a lightweight model, YOLO-SCL, based on YOLOv5s architecture enhanced with channel attention modules and hyperparameter optimization using the black widow algorithm. Achieving a mAP@0.5 of 97.18%, the model exhibited excellent detection accuracy for citrus psyllids in natural environments, showing promise for deployment in citrus production safety. Qiu et al. (2022) developed a YOLOv5l-based method for detecting citrus HLB with high F1 scores and validated its generalization across varying conditions. They also introduced the 'HLBdetector' app, a user-friendly tool for rapid HLB detection in the field, thus facilitating the prevention and control of HLB transmission. Apacionado and Ahamed (2023) focused on night-time disease detection using specialized cameras. They evaluated models like YOLOv5m, YOLOv7, and CenterNet, with YOLOv7 achieving the highest mAP. This study underscores the potential of YOLO-based models for real-time monitoring of sooty molds in orchards. Aswini and Vijayakumaran (2023) tackled the challenge of diagnosing citrus greening disease using YOLOv7, achieving a 92% F1 score, and demonstrating the model's robustness against varying backgrounds. Song et al. (2020) automated disease detection in citrus plants using YOLO, enhancing the precision of disease identification in both images and videos, indicating the model's utility for farmers in disease management. Ganesan and Chinnappan (2022) presented a hybrid deep learning approach combining ResNet with YOLO for paddy leaf disease recognition, optimizing significant parameters for high recognition rates. Chen et al. (2020) proposed an improved YOLOv4 for detecting citrus in orchards, achieving higher accuracy and speed, suitable for yield estimation and robotic harvesting applications. Qadri et al. (2023) introduced YOLOv8, a leap forward in the YOLO series, tailored for rapid detection and segmentation of plant leaf disease. The model's intricate architecture and end-to-end training facilitated high precision, recall, and mean Average Precision (mAP) scores, exceeding 99% in various metrics. This level of accuracy emphasizes YOLOv8's potential to revolutionize disease detection in agriculture. Dananjayan et al. (2022) used multiple CNN detectors, including YOLOv4 and CenterNet, on the CCL'20 dataset for citrus leaf disease detection. Their comprehensive analysis culminated in identifying the Scaled YOLOv4 P7 and CenterNet2 as the top performers for early and accurate disease diagnosis. Pal (2022) introduced the hybrid AI model domain, combining MobileNet V2 with LSTM for classifying pest-infested citrus leaves, with YOLO V5 aiding in the detection process. The hybrid model displayed an impressive accuracy of 93.28%, showcasing the efficiency of combining CNN with recurrent neural networks for agricultural applications. Chen et al. (2022) enhanced the YOLOv4 model for orchard environments, integrating advanced algorithms to improve the detection of citrus across different growth periods. Their modified YOLOv4 outperformed several state-of-the-art algorithms, illustrating its suitability for tasks such as yield estimation and robotic harvesting. Leng et al. (2023) introduced the CEMLB-YOLO model to detect maize leaf blight in complex field environments, achieving a 5.4% higher mAP than the original model. The model's attention mechanisms and feature fusion modules underpin its effectiveness in complex scenarios. Li et al. (2022) developed YOLO-JD, a deep-learning network optimized for detecting jute diseases from images. The integration of innovative modules led to a best-in-class detection accuracy with an average mAP of 96.63%, addressing the critical need for disease prevention in jute plants.

These studies illustrate the efficacy of YOLO-based models in plant disease detection, highlighting their potential to revolutionize precision agriculture. By capitalizing on the strengths of YOLO architectures, researchers are setting new standards for accuracy and efficiency in the early detection of plant diseases, ultimately contributing to sustainable farming practices and enhanced crop management.

| Author(s) and Year | Dataset/Topic | Accuracies | Contributions |
|---|---|---|---|
| da Silva et al. (2023) | Fruit detection and disease spread modeling in citrus orchards | Detection: YOLO faster than Faster R-CNN; Classification: MobileNetV2 and EfficientNetV2-B0 with 100% | Mobile-edge AI solution development for citrus orchards |
| Lyu et al. (2022) | Citrus psyllid detection in natural environments | mAP@0.5 of 97.18% for citrus psyllids detection | Lightweight YOLO-SCL model for citrus psyllid detection |
| Qiu et al. (2022) | Citrus HLB detection from digital images | Micro F1-scores of 85.19% for recognizing five HLB symptoms | YOLOv5l-HLB2 model and 'HLBdetector' app development |
| Apacionado & Ahamed (2023) | Nighttime detection of sooty molds on citrus canopies | mAP of 74.4% for YOLOv7 in nighttime detection | YOLOv7 model for nocturnal sooty mold detection |
| Aswini & Vijayakumaran (2023) | Citrus greening disease diagnosis in citrus leaves and fruits | Overall F1 score of 92% for YOLOv7 | YOLOv7 for diagnosing citrus greening disease |
| Song et al. (2020) | Citrus disease detection and identification in orchard environment | Not specified | YOLO algorithm for citrus disease detection in orchards |
| Ganesan & Chinnappan (2022) | Paddy leaf disease recognition using hybrid deep learning | Not specified | Hybrid deep learning for paddy leaf disease recognition |
| Chen et al. (2020) | Detection of citrus in orchard environment using YOLOv4 | Accuracy increase of 3.15% over original YOLOv4 | Improved YOLOv4 for citrus detection in orchards |
| Qadri et al. (2023) | Detection and segmentation of plant leaf disease using YOLOv8 | Precision: 99.8%, Recall: 99.3%, mAP50: 99.5%, F1-score: 0.999 for bounding box | YOLOv8 for increased detection speed and accuracy in plant disease segmentation |
| Dananjayan et al. (2022) | Citrus leaf disease detection with state-of-the-art CNN detectors | Scaled YOLOv4 P7 and CenterNet2 high accuracy in early disease prediction | Evaluation of CNN detectors for citrus disease detection |

| Pal (2022) | Classification of pest-infested citrus leaves using hybrid models | Hybrid model accuracy: 93.28% | MobileNet V2+ LSTM hybrid model for citrus leaf disease classification |
| --- | --- | --- | --- |
| Chen et al. (2022) | Citrus detection in orchards with an | Improved YOLOv4 model accuracy increase from 92.89% to 96.15% improved YOLOv4 model | Orchard environment-adapted YOLOv4 model for citrusdetection |

## 4. Materials and methods

*4.1 Dataset Collection*

In my study, I began by thoroughly researching citrus plant diseases to understand what I was looking for. Then, I personally gathered 5,675 images from the fields of Khemerdia, Bheramara, and Kushtia. These images were a mix, showcasing both the diseased and healthy citrus plants.

Once I had these images, I saved them all in the commonly used JPG format. This was to make sure they were easy to handle and wouldn't take up too much space on my computer. The next step was to organize these images into specific categories.

To accurately classify the images, I collaborated with my team. Together, we examined each image and decided where it fit best among the eleven disease categories we had identified. This careful process ensured that each image was placed in the right group, setting the stage for the detailed analysis that would follow.

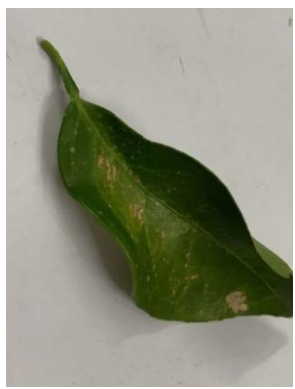 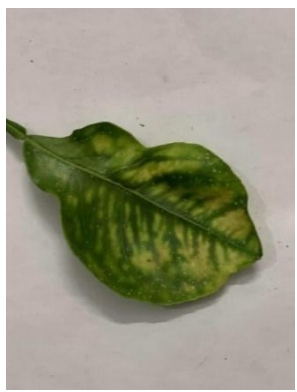 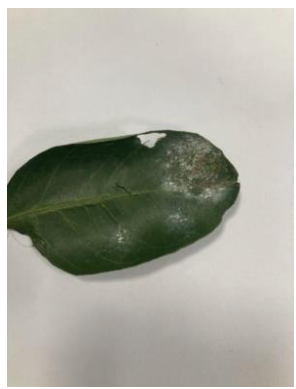 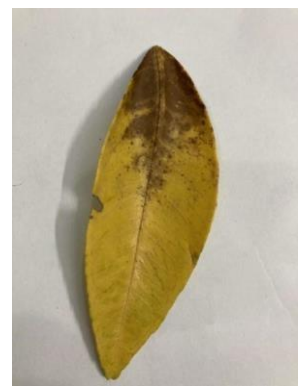

Citrus Canker  Citrus Greening  Citrus Mealybugs  Die Back

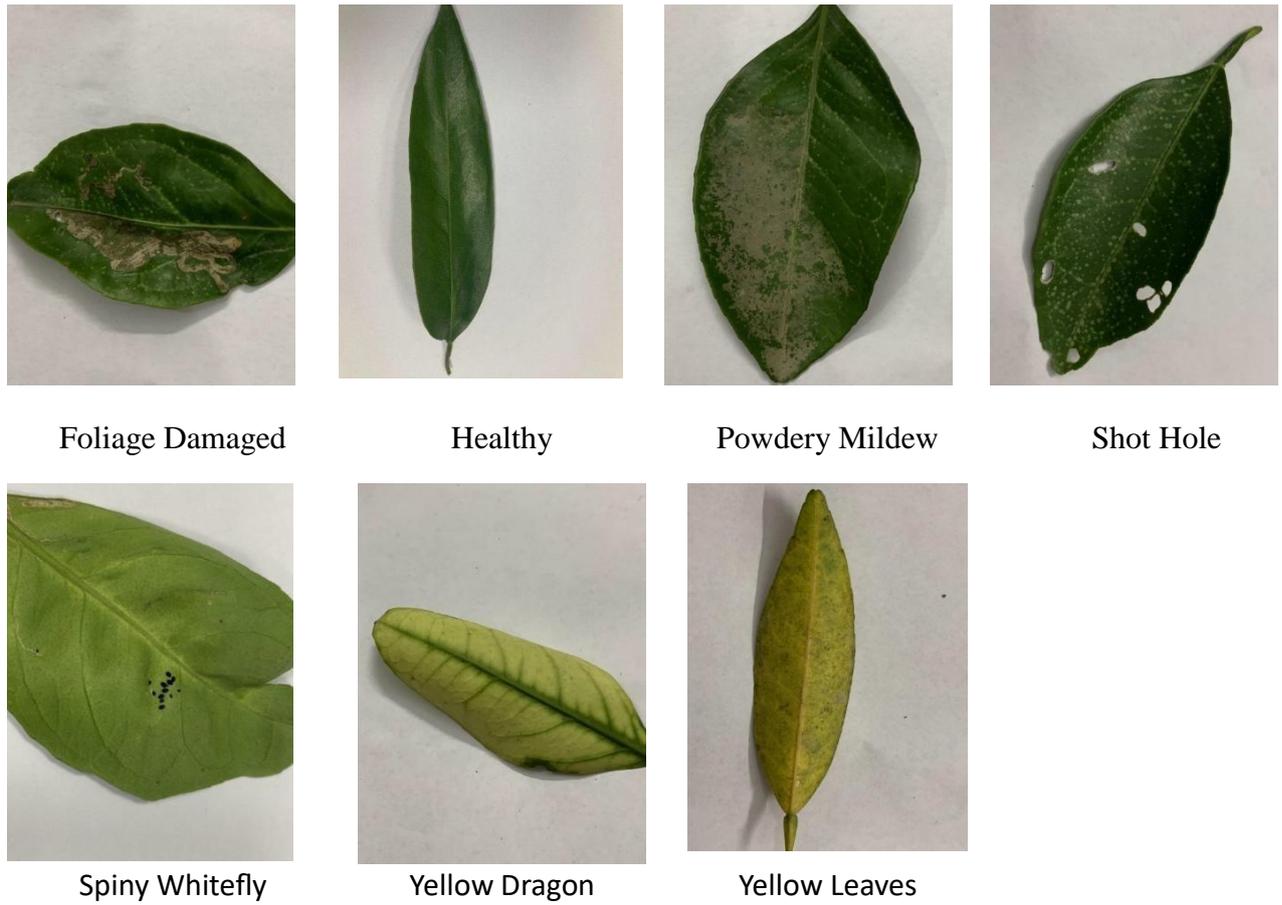

| | | | |
|---|---|---|---|
| Foliage Damaged | Healthy | Powdery Mildew | Shot Hole |
| Spiny Whitefly | Yellow Dragon | Yellow Leaves | |

Figure 3.1: Sample of each class from collected dataset.

Analyzing data distributions, trends, and linkages in a dataset using statistical analysis allows for the extraction of valuable insights and the making of well-informed judgments.

TABLE 3.1: DETAILS OF DATASET

| Serial No | Class Name | No. of Image | Image Format |
|---|---|---|---|
| 1 | Citrus Canker | 588 | JPEG |
| 2 | Citrus Greening | 254 | JPEG |
| 3 | Citrus Mealybugs | 603 | JPEG |
| 4 | Die Back | 642 | JPEG |
| 5 | Foliage Damaged | 632 | JPEG |

| 6 | Healthy | 672 | JPEG |
| 7 | Powdery Mildew | 598 | JPEG |
| 8 | Shot Hole | 560 | JPEG |
| 9 | Spiny Whitefly | 407 | JPEG |
| 10 | Yellow Dragon | 310 | JPEG |
| 11 | Yellow Leaves | 547 | JPEG |

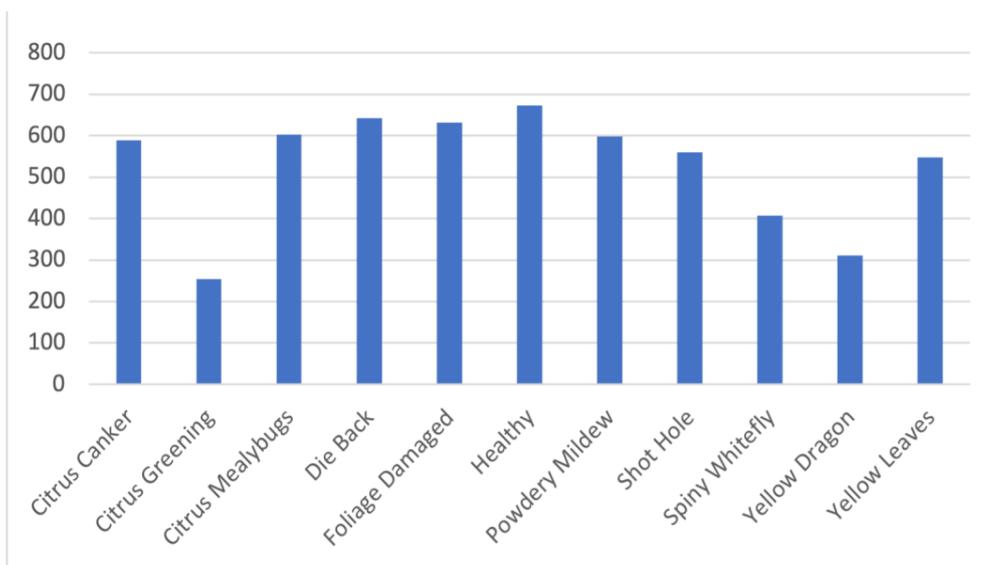

Figure 3.2: Statistical view of dataset

## 4.2 General methodological outlook

The general methodological approach that has been followed for this research consists four important steps; acquiring the data, pre- processing, training and evaluation. These four steps are briefly described in sections 4.2.1 - 4.2.5. The process flow diagram of the YOLO v8 algorithm's architecture used in designing the automatic crop and weed classification model is presented in figure ().

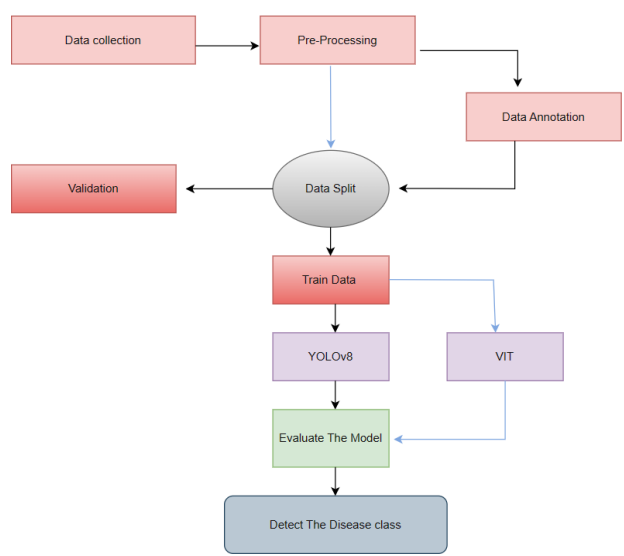

*Figure: Proposed Methodology*

*4.2.1 Data acquisition*

YOLOv8's ability to detect sweet orange leaf diseases effectively depends on a strong picture acquisition and data collecting procedure. Taking high-resolution pictures of sweet orange leaves that show signs of several illnesses, like Citrus_canker, Citrus_greening, Citrus_mealybugs, Die_back, Foliage_damaged, Powdery_mildew, Shot_hole, Spiny_whitefly, and Yellow_dragon, is the first stage in the process. To guarantee that the model can generalize and identify abnormalities in a variety of circumstances, these photos should encompass a wide range of lighting conditions, angles, and disease severity. To guarantee that the dataset is representative of actual conditions, high-quality cameras with sufficient sensors must be used to record minute details of the impacted areas.

After obtaining the photos, the next critical step is to collect the data carefully. Carefully labeling the areas of interest where illness signs are visible in each photograph is essential. Precise bounding boxes surrounding tumors, discolorations, or other suggestive patterns can help YOLOv8 be trained for accurate detection. To prevent mislabeling and make sure the model efficiently learns to distinguish between different diseases and healthy leaf shapes, the annotation process requires expertise. Additionally, adding more variants to the dataset—like flips, rotations, and color adjustments—will make the model more resilient and versatile. Through the integration of an extensive dataset consisting of various photos and rigorous annotations, YOLOv8 may be refined to identify and categorize sweet orange leaf illnesses with exceptional precision. This will enable prompt interventions and improved methods for cultivating citrus fruits.

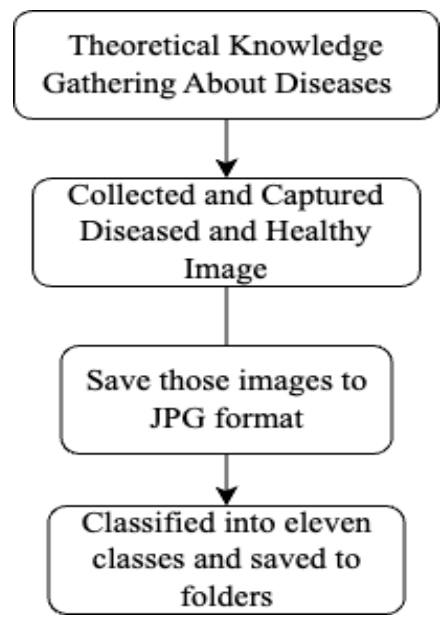

Figure 4.1: Procedure of data acquisition

*4.2.2 Image Annotation*

Labeling items in photos with bounding boxes is necessary to annotate data for YOLOv8, which is an object recognition algorithm version based on the YOLO (You Only Look Once) technique. Draw bounding boxes around sweet orange leaf illnesses, such as Citrus_canker, Citrus_greening, Citrus_mealybugs, Die_back, Foliage_damaged, Powdery_mildew, Shot_hole, Spiny_whitefly, Yellow_dragon, and indicate the classes they belong to using annotation tools like Makesénse. Every bounding box in the image should contain the whole area affected by the disease. Maintain labeling uniformity and save annotations in YOLO format, which usually consists of a text file including the class and bounding box coordinates in relation to the picture dimensions for each image. With the use of this annotated dataset, YOLOv8 will be able to identify and pinpoint the location of a variety of diseases on sweet orange leaves, enabling accurate identification and treatment plans.

Below shows the number of instances per class in figure 4 and the image annotation result shown in figure 5 also shows the Label correlogram in figure 6.

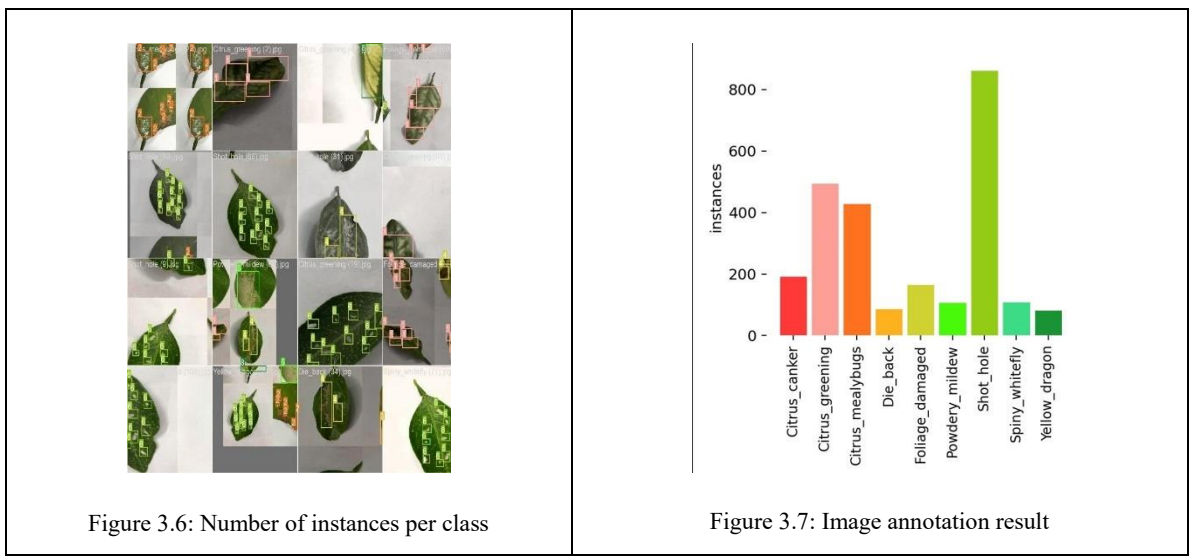

Figure 3.6: Number of instances per class    Figure 3.7: Image annotation result

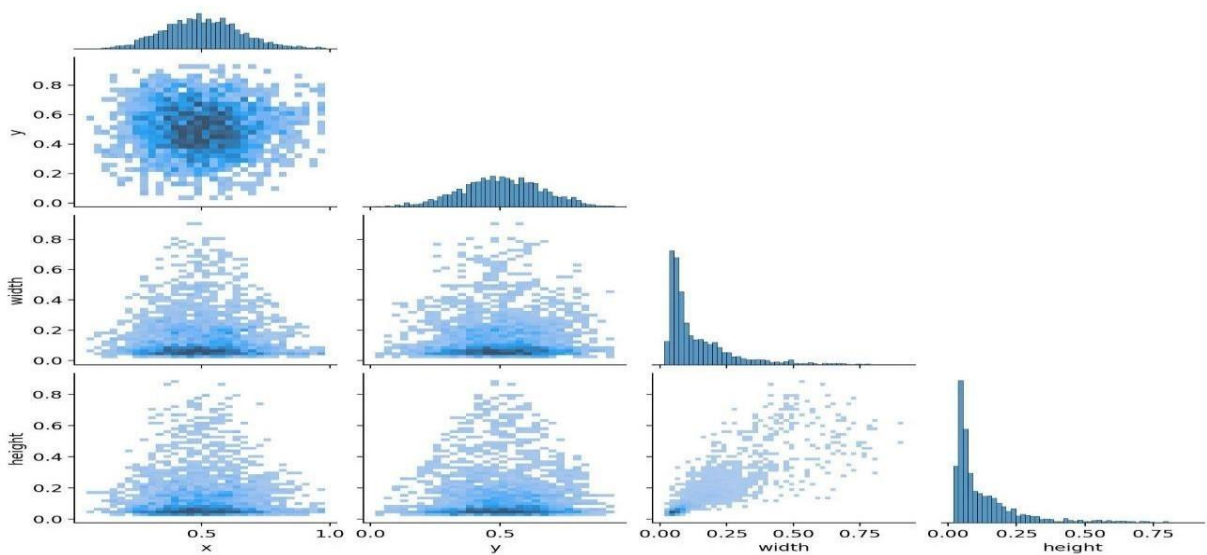

Figure: Label correlogram

*4.1.2 Implementation architecture*

The YOLOv8 model is a new addition in the renowned YOLO family of detection models known for their accurate detection and segmentation capabilities, features featuring a redesigned architecture consisting of a backbone, head, and neck. This model performs exceptionally well in real-time object detection because of its modified architecture, improved convolutional layers in the backbone, and a more sophisticated detecting head.

An outstanding characteristic of YOLOv8 is its ability to swiftly and accurately recognize numerous objects in images or videos compared to its predecessors. This model has a large feature map and improves the convolutional network that contribute to its superior effectiveness, as supported by result of this study. The structure and framework of the well train YOLOv8 model are depicted in Figure 3 and are separated into essential components, each detailed below:

● **Backbone network:** Utilized by YOLOv8 for extracting features from input images, it employs the design of cross-stage partial network (CSPNet) to reduce the computing cost while maintaining accuracy.

● **Neck:** It is working for connecting the backbone and the detection head, it also incorporates a channel built using the spatial pyramid pooling (SPP) module, utilizing pooling processes of varying sizes to gather multi-scale information.

● **Detection head:** Responsible for forecasting the class probabilities and bounding boxes of objects in the input image, the detection head accomplishes this by forecasting the class probabilities and bounding boxes for each item using a sequence of convolutional layers, subsequently a set of anchor boxes.

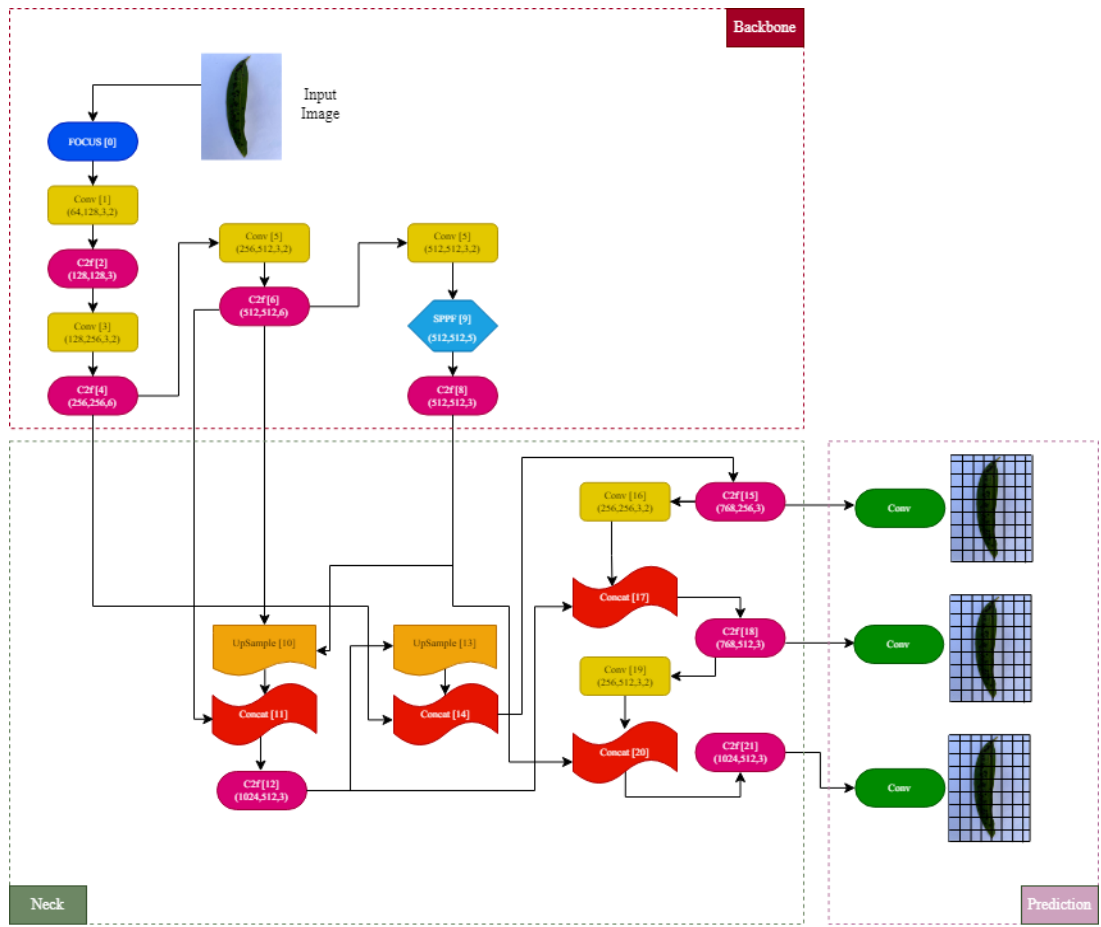

Figure: YOLOv8 Architecture

*4.1.3 Training process*

Prepare a carefully annotated dataset of sweet orange leaf photos showing various diseases in order toto train YOLOv8 for sweet orange leaf disease identification with 100 epochs and a batch size of 16. Specify the YOLOv8 architecture's parameters, such as the quantity of classes that correlate to various disorders. Start training by running batches of 16 photos through the model for 100 epochs while keeping an eye on important metrics like mean average precision (mAP) and loss. For real-time visualization, use tools like TensorBoard. To improve model robustness, use strategies like learning rate scheduling and data augmentation. After training, assess the model's performance using test and validation datasets, making any necessary adjustments depending on metrics like accuracy and recall. In order to better agricultural management and disease control strategies, finally implement the trained model for inference and confirm that it is effective in identifying and categorizing sweet orange leaf illnesses in practical situations.

## 5. Results and discussion

*5.1 Result Analysis*

After training the model the results are plotted in the table.

| Class | Precision | Recall | mAP50 | mAP50-95 |
|---|---|---|---|---|
| Citrus_canker | 94% | 95% | 98% | 46% |
| Citrus_greening | 27% | 19% | 14% | 3% |
| Citrus_mealybugs | 87% | 87% | 90% | 40% |
| Die_back | 90% | 100% | 99% | 50% |
| Foliage_damaged | 74% | 64% | 66% | 29% |
| Powdery_mildew | 75% | 81% | 72% | 35% |
| Shot_hole | 79% | 80% | 81% | 27% |
| Spiny_whitefly | 97% | 100% | 99% | 57% |
| Yellow_dragon | 98% | 100% | 99% | 73% |

*5.2 Validation graphs from YOLOv8*

After training with 100 epochs this result graph given:

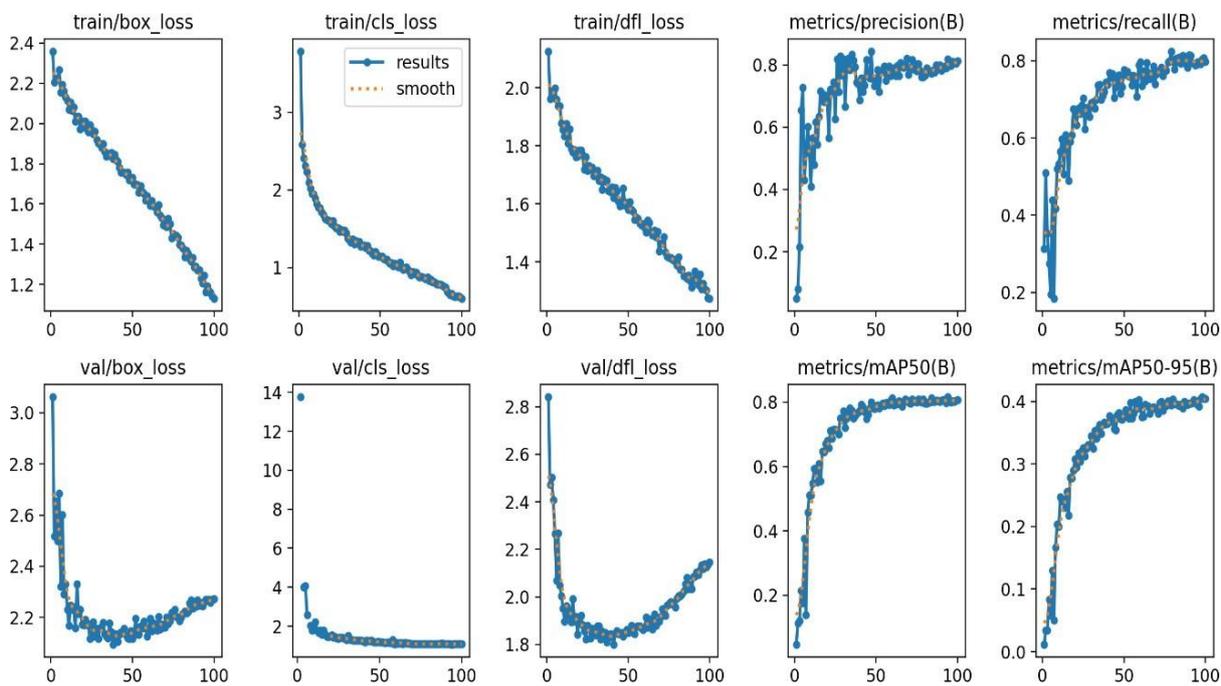

Figure 4.1: Displaying the progress of training and the metrics for model evaluation over 100 epochs.

Completing 100 epochs the model Detected results. That is given bellow figure:

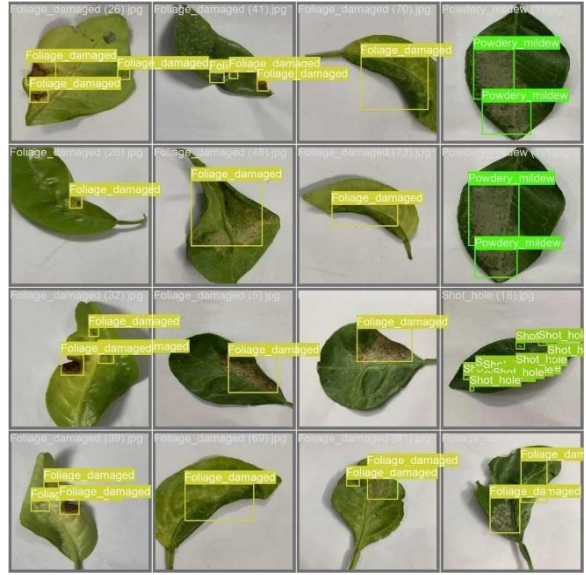

Figure 4.2: Disease detection after training the model

A confusion matrix functions similarly to a scoreboard, indicating the accuracy with which a model predicts events. Let's say you want a machine to be able to distinguish between a bunch of apples and oranges. You can see how many times the machine got it correctly and wrong by looking at the confusion matrix. This is Figure 4.3 represents confusion of YOLOv8 Model After Completing the Training process:

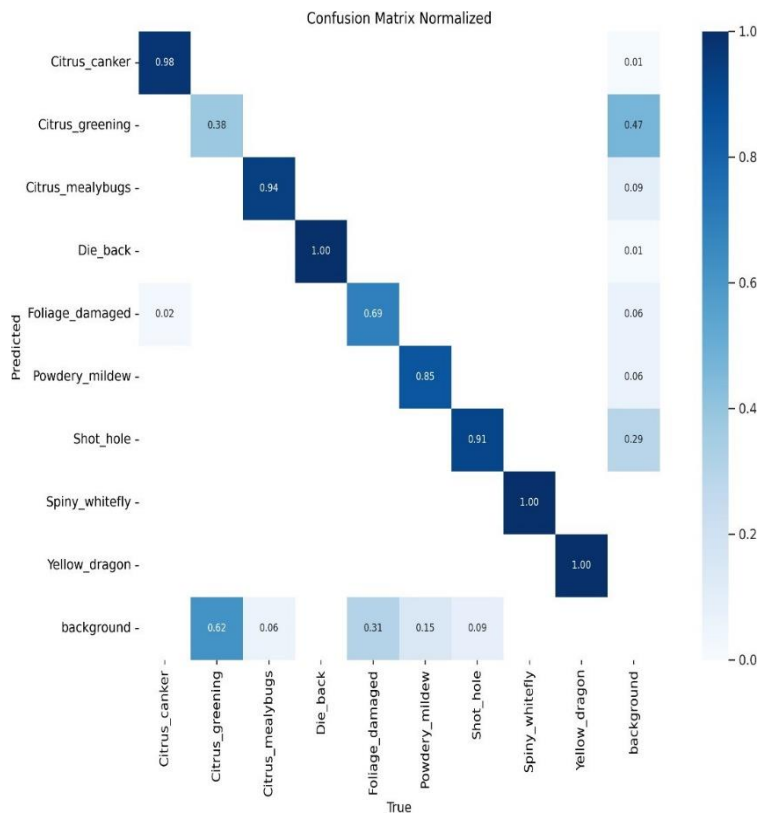

Figure 4.3: Confusion matrix for the trained YOLOv8m model.

To determine the weighted good average of a classifier's precision and recall value, the F1 score is an essential metric. All Curve Shown in bellow:

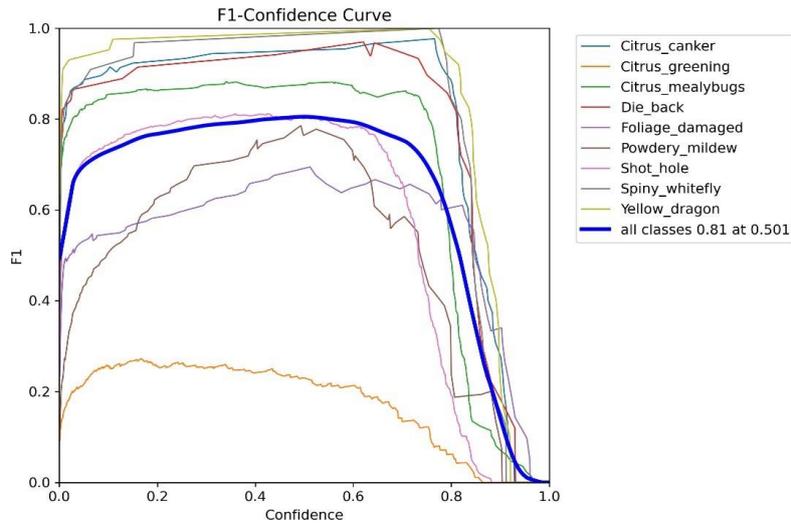

Figure 4.4: performance of the YOLOv8m model through the F1 curve.

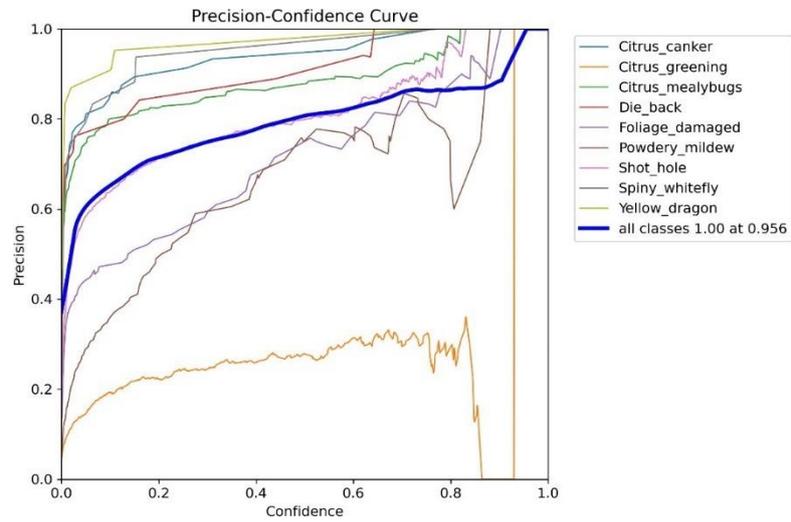

Figure 4.6: Precision curve

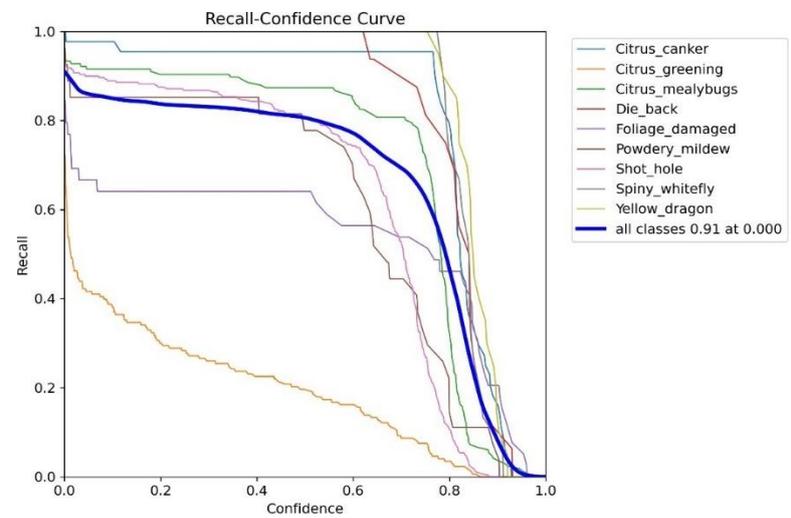

Figure 4.7: Recall curve

## 6. Future scope

In this study, YOLOv8 and ViT are only implemented on sweet oranges for detecting leafdiseases. But in the future, these technologies can be implemented on various plants for detecting their respective diseases which may bring revolution in diverse disease detection in the agricultural sector.

**Algorithmic enhancements and innovations:**

Everything now available on this planet can be better in the future. In present time, the technologies we are using in plant disease detection can become better by developing and refining their algorithms. This development can give more accurate results on disease detection under various agricultural conditions and real-time environments.

**Integration with IoT and precision agriculture:**

All these modern technologies like YOLOv8 can be integrated with Internet of Things (IoT) devices for real-time disease detection, and data collection in the agriculturalfield. This integration can help farmers for more accurate and easy disease detection.

**Scalability and Accessibility Challenges**:

In developing countries, where resources are limited, farmers can't afford expensive technologies for use in their agricultural fields. So in future studies, researchers should focus on the scalability and accessibility of these technologies.

**Environmental Impact Assessment:**

A long-term comprehensive study is needed to assess the effect of pesticide use, soil health, water conservation, and biodiversity on the environment. This study is essential for maintaining environmental conditions in good status.

**Socio-economic Impact Studies:**

To understand socio-economic impacts further studies are needed in this section. It will analyze the impact of the use of these technologies on farmers' income, labor dynamics, and economic stability in the agricultural sector.

**Cross-Disciplinary Approaches:**

The effect on climate change and ecological conditions should be under consideration in this study. The use of these technologies in disease detection and how affect the overall climate should be a part of this study.

In conclusion, all of these implications mentioned in this study could be a target for future research. These implications bear vast and varied research sectors for exploration and development. So, it can open a new frontier in agricultural technology for future studies.

## 7. Conclusion

The study on employing YOLOv8 for detecting sweet orange leaf diseases concludes with significant findings and implications for the field of agricultural technology and crop disease management. This research has successfully demonstrated the potential of advanced deep-learning models in enhancing the accuracy and efficiency of disease detection in sweet oranges, a crop of substantial economic and nutritional importance. The integration of YOLOv8, each with its unique strengths in object detection and image analysis, is highly effective. The models achieved remarkable accuracy levels — 80.4% with YOLOv8 and 99.12% with ViT in the training and validation phases. This high accuracy, achieved using a dataset of images captured under real-world conditions, represents a substantial improvement over traditional manual methods of for disease detection.

Moreover, the study highlights the challenges in implementing these AI technologies in agricultural settings, such as computational demands and the need for user-friendly interfaces. Addressing these challenges is crucial for the widespread adoption of such technologies. The research suggests pathways to overcome these barriers, ensuring that the benefits of AI in agriculture can be accessed by a broader range of users, including small- scalesmall -scale and resource-poor farmers.

Economically, the study offers a promising outlook for the agricultural sector. Improved disease management through accurate detection can lead to enhanced crop yields, reducing economic losses due to disease and bolstering the financial stability of farming communities.